\documentclass{article} %
\usepackage{iclr2026_conference,times}

\usepackage{amsmath,amsfonts,bm}

\def\eqref#1{equation~\ref{#1}}

\def\1{\bm{1}}

\DeclareMathAlphabet{\mathsfit}{\encodingdefault}{\sfdefault}{m}{sl}
\SetMathAlphabet{\mathsfit}{bold}{\encodingdefault}{\sfdefault}{bx}{n}

\usepackage[hidelinks]{hyperref} %
\usepackage{url}
\usepackage{graphicx}
\usepackage{float}
\usepackage{subcaption}

\title{Latent Perspective‑Taking \\ via a Schr\"odinger Bridge \\ in Influence‑Augmented Local Models}

\author{Kevin Alcedo~\textsuperscript{1,2}, Pedro U. Lima~\textsuperscript{1,2}, Rachid Alami~\textsuperscript{3,4}\\
\textsuperscript{1}Institute for Systems and Robotics, \\
\textsuperscript{2}Instituto Superior Técnico, Universidade de  Lisboa (Lisbon ELLIS Unit), 
\\
\textsuperscript{3}LAAS-CNRS, \\
\textsuperscript{4}Artificial and Natural Intelligence Toulouse Institute (ANITI) \\
\fontsize{8}{10}\texttt{\{kevin.alcedo,pedro.lima\}@tecnico.ulisboa.pt},
\fontsize{8}{10}\texttt{rachid.alami@laas.fr}\\
}

\iclrfinalcopy %
\begin{document}

\maketitle

\begin{abstract}
Operating in environments alongside humans requires robots to make decisions under uncertainty. In addition to exogenous dynamics, they must reason over others' hidden \textit{mental-models} and \textit{mental-states}. While Interactive POMDPs and Bayesian Theory of Mind formulations are principled, exact nested-belief inference is intractable, and hand-specified models are brittle in open-world settings. We address both by learning structured mental-models and an estimator of others’ mental-states. Building on the Influence-Based Abstraction, we instantiate an Influence-Augmented Local Model to decompose socially-aware robot tasks into local dynamics, social influences, and exogenous factors. We propose (a) a neuro-symbolic world model instantiating a factored, discrete Dynamic Bayesian Network, and (b) a perspective-shift operator modeled as an amortized Schr\"odinger Bridge over the learned local dynamics that transports factored egocentric beliefs into other-centric beliefs. We show that this architecture enables agents to synthesize socially-aware policies in model-based reinforcement learning, via decision‑time mental‑state planning (a Schr\"odinger‑Bridge in belief space), with preliminary results in a MiniGrid social navigation task.
\end{abstract}

\section{Introduction}
A socially-aware robot acts alongside humans and robots, pursuing its own goals while enabling others to achieve theirs and accommodating known or inferred needs. Hence, it must make decisions under partial observability: both of the environment and of others' \emph{mental-models} and \emph{mental-states}. We adopt the lens of classical Interactive POMDPs (I-POMDPs)~\citep{2004IPOMDP} and Bayesian Theory of Mind (BToM)~\citep{BAYESIANINVPLAN_BAKER2011}: \emph{mental-models} are (known or inferred) internal transition, observation, reward and action models, and \emph{mental-states} are probability distributions over latent states (beliefs). The knowledge-representation community has formalized perspective-taking~\citep{1992PerspectivesOnPerspectiveTaking} with ontologies/semantic memory~\citep{PerspectiveTakingAlami,2019Ontologenius}, epistemic human-aware task planners~\citep{2022HATPEHDA}, and dynamic epistemic logic~\citep{Bolander2011EpistemicPF,2021Bolander}, yielding interpretable and verifiable systems. Yet these symbolic models are brittle in open-world settings with stochastic dynamics, and exact nested-belief reasoning quickly becomes intractable; PSPACE at level-0, and EXPSPACE in the worst case with factored dynamics and exponential horizon~\citep{2000ComplexityMDPs}. In contrast, probabilistic sequential decision-making naturally handles uncertainty but typically assumes prescribed models and still suffers from the nested-belief explosion. 

As a step toward unifying symbolic knowledge representation and probabilistic sequential decision-making, we propose a data-driven architecture that learns (i) factored, discrete mental-models and (ii) a tractable belief-to-belief map for perspective-taking in factored latent space. The key idea is to cast perspective-taking as belief transport via a Schr\"odinger Bridge (SB)~\citep{1940Fortet_SB} defined over learned local reference dynamics, producing other-centric beliefs compatible with local dynamics. Building on the Influence-Based Abstraction (IBA) framework~\citep{oliehoek2021}, we instantiate an Influence-Augmented Local Model (IALM) to decompose socially-aware robot tasks into local dynamics, social influence, and exogenous factors. We operationalize this with a neuro-symbolic world model and a perspective-shift operator as an amortized SB that runs online at decision-time. Inspired by Simulation Theory~\citep{GOLDMAN1992}, the agent seeds a mental-state and reuses its cognitive machinery to simulate others' mental-state dynamics, yielding a \emph{``what-if''} other-centric belief for decision-making akin to model predictive control in mental-state space. 

Preliminary results in a grid-world social navigation task show our SB-based perspective-taking learns faster and achieves higher returns than using the true context-free current belief. Our contributions are: (1) a practical influence-augmented decomposition for social tasks; (2) a factored, discrete neuro-symbolic world model; and (3) perspective-taking via an amortized Schr\"odinger Bridge that enables model-based belief-space planning at decision-time.

\begin{figure*}[t]
    \centering
    \hfill
    \begin{minipage}[c]{0.25\textwidth}
        \centering
        \subfloat[]{\includegraphics[height=3.5cm, trim=2mm 2mm 2mm 2mm,clip]{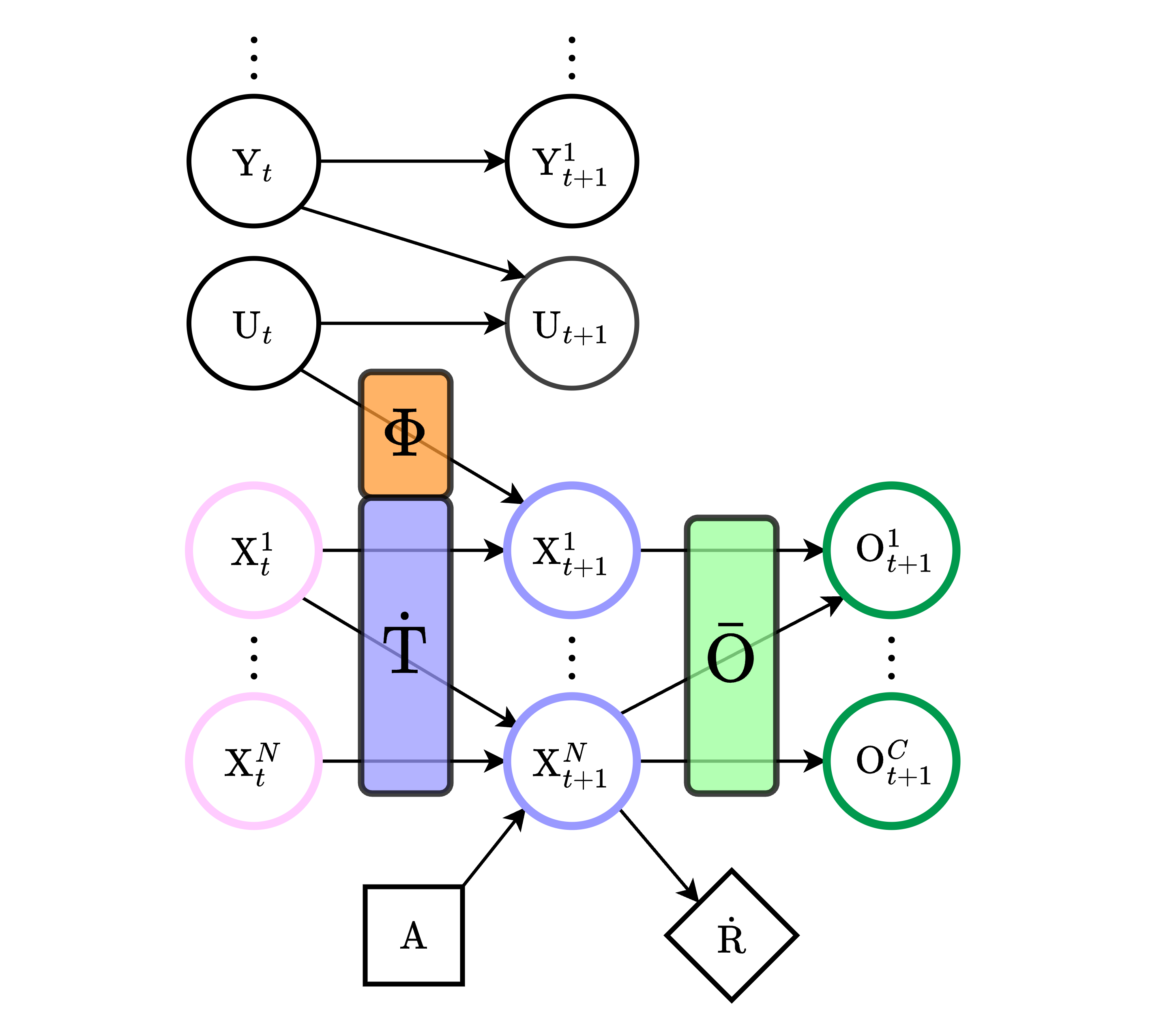}\label{fig:dbn}}
    \end{minipage}
    \hfill
    \begin{minipage}[c]{0.7\textwidth}
        \centering
        \subfloat[]{\includegraphics[height=3.5cm, trim=2mm 2mm 2mm 2mm,clip]{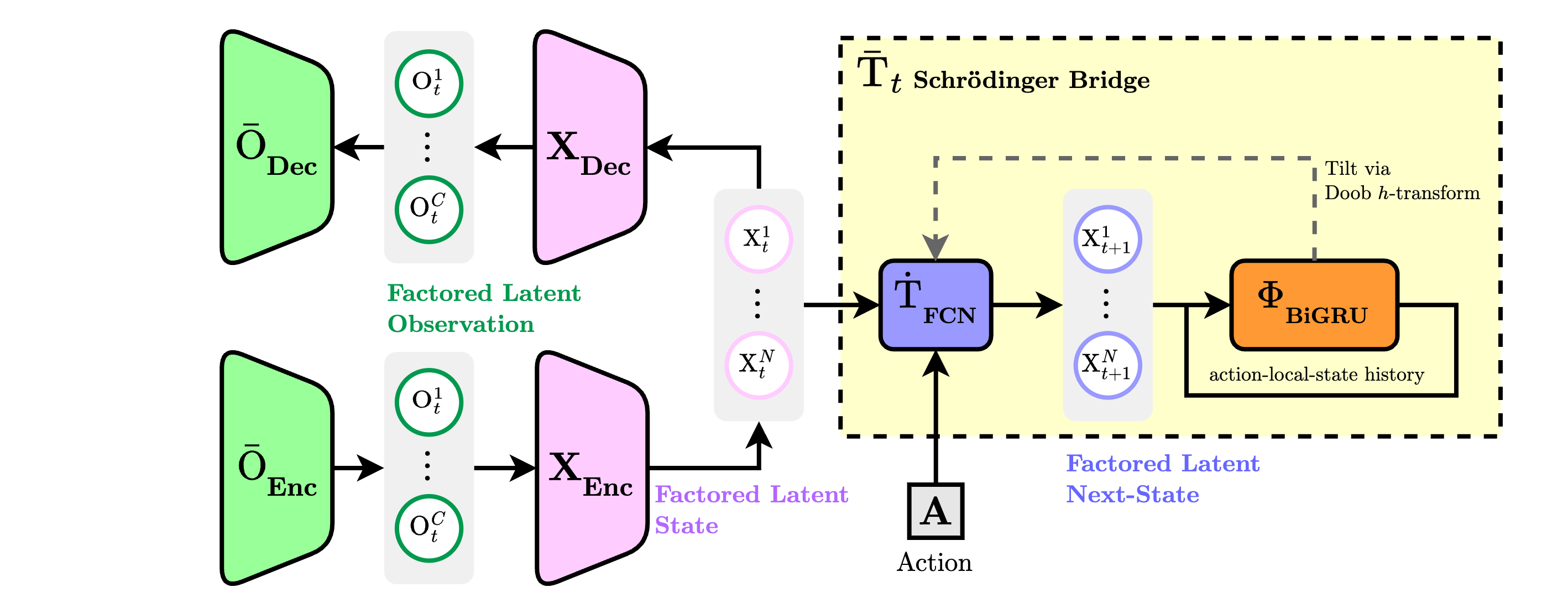}\label{fig:architecture}}
    \end{minipage}
    \hfill
    \caption{(a) Influence-Augmented Local Model; (b) Neuro-Symbolic World Model Architecture.}
    \label{fig:pgm_architecture}
\end{figure*}

\section{Background}
\subsection{Influence-Augmented Local Model}
The IBA framework~\citep{oliehoek2021} addresses the challenge of partitioning large state spaces in structured multi-agent settings by factorizing the global model into smaller, tractable local models and by identifying influence points that compactly capture boundary effects. When modeled exactly, the influence is a sufficient statistic that enables an IALM to compute an exact best response (for fixed external policies) while avoiding reasoning over the full global space. In practice, enumerating all influence sources and updating their estimates is intractable, and approximations are used~\citep{suau22a_ials}. The factored representation makes the probabilistic dependency structure explicit and yields a compact two-stage Dynamic Bayesian Network (2DBN)~\citep{Boutilier1999_2DBN,Poupart_FPOMDP}.

An IALM extends a local factored POMDP (FPOMDP) with a factored global state $\mathcal{S}=\{S_1,\dots,S_k\}$ partitioned into the following variables: local state  $\mathcal{X}$, influence sources $\mathcal{U}$, and non-local states $\mathcal{Y}$. Let   \mbox{$s_t \in S = \times_{i=1}^k S_i$}, \mbox{$x_t \in X = \times_{i\in\mathcal{X}} S_i$},  \mbox{$u_t \in U = \times_{i\in\mathcal{U}} S_i$}, and action $a_t \in \mathcal{A}$ at time $t$. The key insight from~\cite{oliehoek2021} is that a minimal $d$-separating set~\citep{2009KollerFriedman} $D_{t}$, a subset of the action-local-state history \mbox{$h_t = \langle x_1, a_1, \dots, a_{t-1}, x_t\rangle$}, is sufficient to predict the influence distribution \mbox{\(I(u_{t} \mid D_{t})\)}. $D_{t+1}$ is constructed deterministically with function $d(\cdot)$: $D_{t+1}=d(h_t)$. We arrive at the tuple \mbox{$\mathcal{M}^{\mathrm{IALM}} = \langle \bar{\mathcal{S}},\mathcal{A}, \bar{T}, R, \mathcal{O}, \bar{O}, b_0\rangle$}, where $\bar{\mathcal{S}}$ is the augmented local state space \mbox{$\bar{s}_t = \langle x_t, D_{t+1} \rangle$}, \mbox{$\bar{T}(x_{t+1} \mid \bar{s}_t, a_t)$} is the augmented local transition, \mbox{$\bar{O}(o_{t+1} \mid  a_t, x_{t+1})$} is the local observation with $o_{t+1} \in \mathcal{O}$, \mbox{$R(x_t, a_t, x_{t+1})$} is the local reward, and $b_0$ is an initial local belief. The influence adjusts the local transition by marginalization: 
{\small \begin{equation} \label{eq:IALM_T}
\bar{T}(x_{t+1}\mid\bar{s}_t,a_t)
=\sum_{u_{t+1}\in U} \Pr(x_{t+1}\mid x_t,a_t,u_{t+1})\; I(u_{t+1}\mid D_{t+1})
\end{equation}}

\subsection{Schr\"odinger Bridge}

Originally conceived to explain the most likely path distribution (the bridge) between the start and end points of moving particles under an assumed dynamics model, a Schr\"odinger Bridge (SB) can be formulated as the probability law on paths linking two endpoint marginals~\citep{Schrodinger1931,2021_translation_Schrodinger1931}; the goal is to discover a new law describing the dynamics of the trajectory. SB can be viewed as a special case of Optimal Transport: \emph{dynamic entropic optimal transport with reference dynamics}~\citep{2014_Leonard}; we minimize a pathwise KL divergence to the trajectory distribution induced by the reference transition model.

Given a reference local dynamics \mbox{$\dot{T}(x_{t+1}\mid x_t,a_t)$}, start/end marginals (beliefs) $b_0$ and $b_n$, and conditioning on a fixed action sequence $a_{0:n-1}$, we seek a bridge \mbox{$\bar{T}=\{\bar{T}_t\}^{n-1}_{t=0}$} (a family of time-varying transitions) whose induced trajectory distribution is closest to that of the reference $\dot{T}$. Concretely, we compute $\bar{T}$ such that its induced trajectory distribution $\mathbb{P}_{\bar{T}}$ minimizes $\mathrm{KL}\big(\mathbb{P}_{\bar{T}}\,\|\,\mathbb{P}_{\dot{T}}\big)$, subject to the endpoint marginals ($b_0,b_n$). This yields a first-order, time-inhomogeneous Markov chain. Here we focus on the space-time Sinkhorn method~\citep{1940Fortet_SB}, where  forward and backward potentials $\phi_t, \psi_t \in \mathbb{R}_{>0}^{|\mathcal{X}|}$ are propagated through the reference $\dot{T}$ and produce the bridge. We parametrize the potentials with context $c_t := D_{t+1} = d(h_t)$ with $h_t = \langle x_1, a_1, \dots, a_{t-1}, x_t\rangle$. The Schr\"odinger system becomes:
{\small\begin{equation} \label{eq:SB_system}
\begin{aligned}
\psi_t(x_t;c_t) = \sum_{x_{t+1}}\dot{T}(x_{t+1} \mid x_t, a_t)\psi_{t+1}(x_{t+1};c_{t+1}), \\
\phi_{t+1}(x_{t+1};c_{t+1}) = \sum_{x_{t}} \phi_{t}(x_{t};c_t) \dot{T}(x_{t+1} \mid x_t, a_t), \\ 
p_t(x_t) = \phi_t(x_t;c_t)\psi_t(x_t;c_t), \quad p_0=b_0, \;p_n =b_n
\end{aligned}
\end{equation}}

The steered transitions take the Doob $h$-transform form~\citep{1957Doob}: 
{\small \begin{equation} \label{eq:SB_doob}
\bar{T}_t(x_{t+1}\mid \bar{s}_t,a_t)
= \dot{T}(x_{t+1}\mid x_t,a_t)\frac{\psi_{t+1}(x_{t+1}; c_{t+1})}{\psi_{t}(x_t; c_{t})} , \quad \mathrm{where} \; \bar{s}_t :=\langle x_t, D_{t+1} \rangle
\end{equation}}
The induced trajectory distribution, with initial state distribution $\mu_0$, is:
{\small \begin{equation} \label{eq:SB_path}
\mathbb{P}_{\bar{T}}(x_{0:n}) \propto \phi_0(x_0; c_{0}) \mu_0(x_0) \bigg \lbrack \prod_{t=0}^{n-1} \dot{T}(x_{t+1} \mid x_t, a_t) \bigg \rbrack \psi_n(x_n; c_n)
\end{equation}}

One way to view this process is as control-as-inference~\citep{NIPS2006_Todorov}, targeting a minimum-effort steering of a noisy passive dynamic process. In Section~\ref{sec:Method} we instantiate an IALM as a neuro-symbolic world model that learns $(\bar{\mathcal{S}}, \dot{T}, \mathcal{O}, \bar{O}, D)$. We then realize $\bar{T}=\{\bar{T}_t\}$ as an approximation of the ideal IALM transition in Eq.~\ref{eq:IALM_T} via an amortized Schr\"odinger Bridge tilt of $\dot{T}$ for perspective-taking. The potentials $\phi$ and $\psi$ are trained offline with epistemic counterfactuals. 

\section{Method}
\label{sec:Method}
\subsection{Neuro-Symbolic World Model}
We instantiate the IALM as a neuro-symbolic world model, composed of a hierarchy of discrete variational-autoencoders (dVAEs) with Gumbel-Softmax~\citep{GumbelSoftmax_01, GumbelSoftmax_02} trained end-to-end on transitions $(o_t,a_t,o_{t+1})$, explicitly modeling the structural components of a 2DBN (See Fig.~\ref{fig:pgm_architecture}). Raw observations are encoded into a latent of shape $(C,F)$, where $C$ is the number of discrete observation factors and each has $F$ categories. These are then mapped to factored state latents $(N,K)$, with $N$ state factors each with $K$ categories. A local, influence-na\"ive transition model $\dot{T}$ maps the current factored state and a one-hot action to the next factored latent state. The transition model uses $\alpha$-entmax~\citep{2019Peters_alphaentmax} cross-attention over the input state factors to induce exact-zero, data-driven sparsity, exploiting latent parent–child structure.

\subsection{Perspective-Shift Operator}
Extending~\citet{2025Alcedo_ROMAN}, we estimate others' beliefs via an SB-based perspective-shift operator: $\hat{b}_{i} = \Phi_{\mathrm{SB}}(b_0, \hat{x}_{i}, a_{0:n-1}, \dot{T})$, where $b_0$ is the egocentric belief, $\hat{x}_{i}$ is the locally observed relative state of another agent $i$, $\dot{T}$ is the learned local influence-na\"ive transition model (the reference dynamics), $a_{0:n-1}$ is a proposed action sequence hypothesized for agent $i$ generated via an upstream local planner, and $\hat{b}_{i}$ is the estimated other-centric belief. We implement $\Phi_{\mathrm{SB}}$ by learning the potentials ($\hat{\phi_t}, \hat{\psi_t}$) using a bi-directional sequence model (BiGRU) with $\alpha$-entmax attention on the action-local-state history \mbox{$h_t = \langle x_1, a_1, \dots, a_{t-1}, x_t\rangle$} generated from a roll-out using the reference model $\dot{T}$. The sparsity induced by $\alpha$-entmax on $h_t$ serves as the deterministic update to compute the approximate minimal d-set from the action-local-state history $D_{t+1}=d(h_t)$. At training time, we sample  ``epistemic counterfactuals'' (other-centric beliefs) $b_n$ in a single-agent task from the environment to amortize the potentials that realize $\bar{T}$ via the Doob $h$-transform of $\dot{T}$, minimizing $\mathrm{KL}\big(\mathbb{P}_{\bar{T}}\,\|\,\mathbb{P}_{\dot{T}}\big)$ subject to the endpoints. At inference, we tilt $\dot{T}$ with the amortized potentials ($\hat{\phi_t}, \hat{\psi_t}$) parametrized by $c_t := D_{t+1} = d(h_t)$ to reach an estimate for $\hat{b}_i$. The learning objective is done at the per-factor level, enabling $d(h_t)$ to contextually manage the tilt by masking factors. The objective becomes: $\mathcal{L}=\mathcal{L}_{\bar{O}} + \mathcal{L}_{\dot{T}} + \mathcal{L}_{\Phi}$. We employ training stabilization techniques that will be discussed and ablated in the full manuscript; these include: Gumbel-softmax $\tau$ annealing, knowledge distillation confidence-gating, and socially-weighted epistemic counterfactuals objectives at two levels of reasoning. 

\begin{figure*}[t]
    \centering
    \hfill
    \begin{minipage}[c]{0.4\textwidth}
        \centering
        \subfloat[]{\includegraphics[height=3.2cm, trim=2mm 2mm 2mm 2mm,clip]{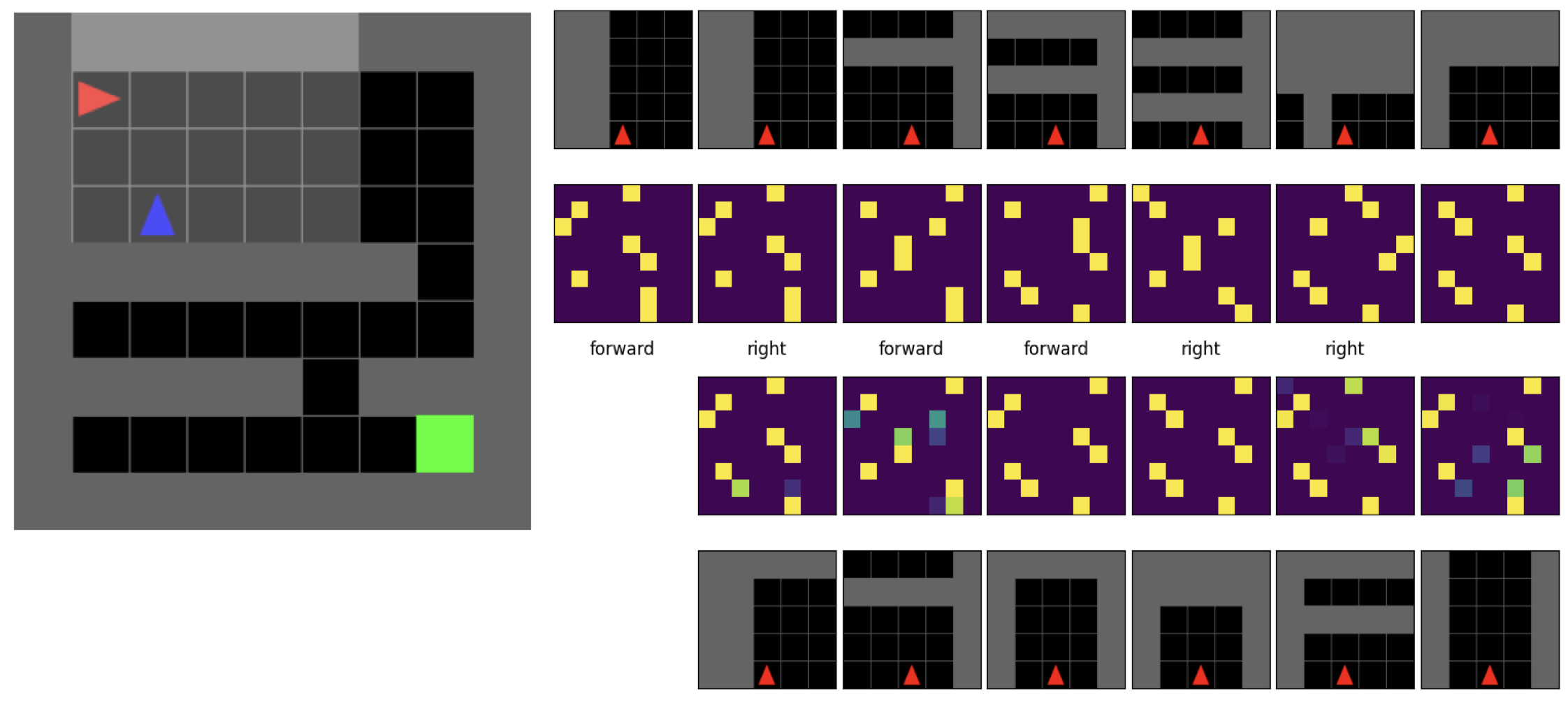}\label{fig:results_a}}
    \end{minipage}
    \hfill
    \begin{minipage}[c]{0.5\textwidth}
        \centering
        \subfloat[]{\includegraphics[height=3.5cm, trim=2mm 2mm 2mm 2mm,clip]{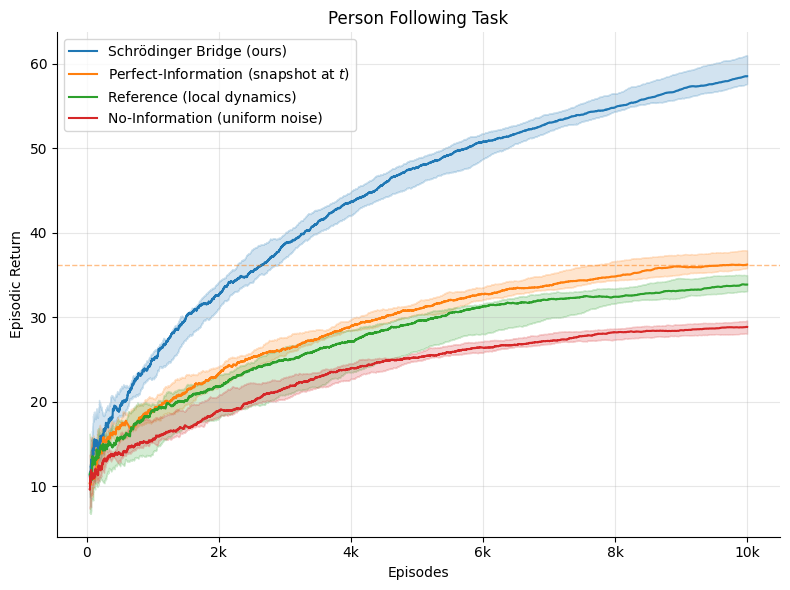}\label{fig:results_b}}
    \end{minipage}
    \hfill
    \caption{(a) Perspective-shift representations. Left: global state. Right: ground-truth observations (top); factored, discrete latent states (second); actions (third); next-step latent states (fourth); reconstructed observation (last). (b) Median episodic returns over 5 seeds, shaded band denotes 95\% bootstrap CI of the median; dashed line: Perfect-Information baseline final median.}
    \label{fig:results}
\end{figure*}

\section{Preliminary Results, Limitations \& Next Steps}
We evaluate our approach for influence-aware policy learning in a multi-agent model-based reinforcement learning setting. \textbf{Environment:} partially-observable Minigrid~\citep{MinigridMiniworld23}, with discrete observations and actions. \textbf{Person-following Task:} follow a fixed‑policy agent to an unknown target while satisfying social norms~\citep{2025Alcedo_ROMAN}. The environment rewards proximity and visibility, gives a large terminal reward for reaching the goal, penalizes collisions, and truncates episodes when the follower stays too far away for too long. \textbf{Influence-aware Policy:} $a_t = \pi_{\mathrm{DQN}}(b^0_t,\hat{b}^{i}_t)$. \textbf{Baselines:} (i) Perfect-information, true current belief of the other agent with no context/history: $\hat{b}^{i}_t=b^i_t$, (ii) No-information: $\hat{b}^{i}_t \sim \mathrm{Uniform}(\Delta_X)$, and (iii) Reference (local dynamics) multi-step rollout: $\hat{b}^{i}_t = \prod^{n-1} \dot{T}(\cdot)$. \textbf{Results:} SB learns faster and achieves higher returns than all baselines (See Fig.~\ref{fig:results}), highlighting the benefit of decision‑time, context‑aware mental‑state planning over context‑free estimates. Currently, our approach assumes homogeneity in the perception and action models of other agents, we aim to address this in future work by adapting the belief-to-belief transform objective via heterogeneous epistemic counterfactuals.  
\section{Related Work}
Latent perspective‑taking has been approached via meta‑learned agent embeddings~\citep{2018Rabinowitz_ToM}, latent state‑space world models that generate other‑centric observations~\citep{2023LEAPT}, and latent opponent/teammate modeling under partial observability \citep{2020Labash_neuro_perspective-taking,2021Papoudakis}. In contrast, we cast perspective‑taking as belief‑to‑belief transport, implementing an amortized SB within a learned IALM to produce other‑centric beliefs for decision‑time mental-state planning; we extend the local-model-only approach of~\citet{2025Alcedo_ROMAN}.~\citet{suau22a_ials} instantiate IALMs at scale in multi-agent reinforcement learning settings, with prescribed influence sources and learned predictors, and ~\citet{Suau2025_IAM} learn influence‑aware memory and a d-set selector in pixel space (ours operates in latent space). Finally, \citet{2023generalized_belief_transport} proposed the generalized belief transport method for teacher-student learning settings modeled via a static unbalanced entropic optimal transport, whereas we use a Schr\"odinger Bridge.

\subsubsection*{Acknowledgments}
This work was supported by LARSyS FCT funding (DOI: 10.54499/LA/P/0083/2020, 10.54499/UIDP/50009/2020, and 10.54499/UIDB/50009/2020) and the Portuguese Recovery and Resilience Plan (PRR) through project C645008882-00000055, Center for Responsible AI; and partially supported by ANR (Agence Nationale de la Recherche) AAPG- 2024/Ostensive project; and the project European Learning and Intelligent Systems Excellence (ELIAS, Grant Agreement No. 101120237). 

\bibliography{iclr2026_conference}
\bibliographystyle{iclr2026_conference}

\end{document}